\documentclass{article}

\usepackage{PRIMEarxiv}

\usepackage[utf8]{inputenc} 
\usepackage[T1]{fontenc}    
\usepackage{hyperref}       
\usepackage{url}            
\usepackage{booktabs}       
\usepackage{amsfonts}       
\usepackage{nicefrac}       
\usepackage{microtype}      
\usepackage{lipsum}
\usepackage{fancyhdr}       
\usepackage{graphicx}       
\graphicspath{{media/}}     
\usepackage{multirow}
\usepackage{color}
\usepackage[normalem]{ulem}
\useunder{\uline}{\ul}{}
\usepackage{array} 
\usepackage{xcolor}
\usepackage{colortbl}
\usepackage{float}

\let\ul\undefined 
\usepackage[normalem]{ulem} 
\usepackage{soul} 
\usepackage{stfloats}

\title{Integrated Framework for LLM Evaluation with Answer Generation} 

\author{
Sujeong Lee\thanks{These authors contributed equally to this work; the order of authorship was determined by random draw.} \\
Inha University\\
Incheon, 22212, Republic of Korea \\
\texttt{tnwjd025611@inha.edu} 
\And
Hayoung Lee\footnotemark[1] \\
Inha University\\
Incheon, 22212, Republic of Korea \\
\texttt{gkdud000123@gmail.com} 
\And
Seongsoo Heo \\
Inha University\\
Incheon, 22212, Republic of Korea \\
\texttt{woo555813@inha.edu} 
\And
Wonik Choi \\
Inha University\\
Incheon, 22212, Republic of Korea \\
\texttt{wichoi@inha.ac.kr} 
}

\begin{document}
\maketitle

\begin{abstract}
Reliable evaluation of large language models is essential to ensure their applicability in practical scenarios. Traditional benchmark-based evaluation methods often rely on fixed reference answers, limiting their ability to capture important qualitative aspects of generated responses. To address these shortcomings, we propose an integrated evaluation framework called \textit{self-refining descriptive evaluation with expert-driven diagnostics}, SPEED, which utilizes specialized functional experts to perform comprehensive, descriptive analyses of model outputs. Unlike conventional approaches, SPEED actively incorporates expert feedback across multiple dimensions, including hallucination detection, toxicity assessment, and lexical-contextual appropriateness. Experimental results demonstrate that SPEED achieves robust and consistent evaluation performance across diverse domains and datasets. Additionally, by employing relatively compact expert models, SPEED demonstrates superior resource efficiency compared to larger-scale evaluators. These findings illustrate that SPEED significantly enhances fairness and interpretability in LLM evaluations, offering a promising alternative to existing evaluation methodologies. 
\end{abstract}


\section{Introduction}
In recent years, rapid advancements in \textit{large language models} (LLMs) have profoundly impacted the field of \textit{artificial intelligence} (AI) and various industries, including healthcare, law, and finance. These breakthroughs have accelerated the adoption of LLMs across diverse domains, underscoring the need for a reliable evaluation framework capable of systematically assessing their strengths and limitations~\cite{guo2023evaluatinglargelanguagemodels}.

\begin{figure}
    \centering
    \includegraphics[width=0.5\linewidth]{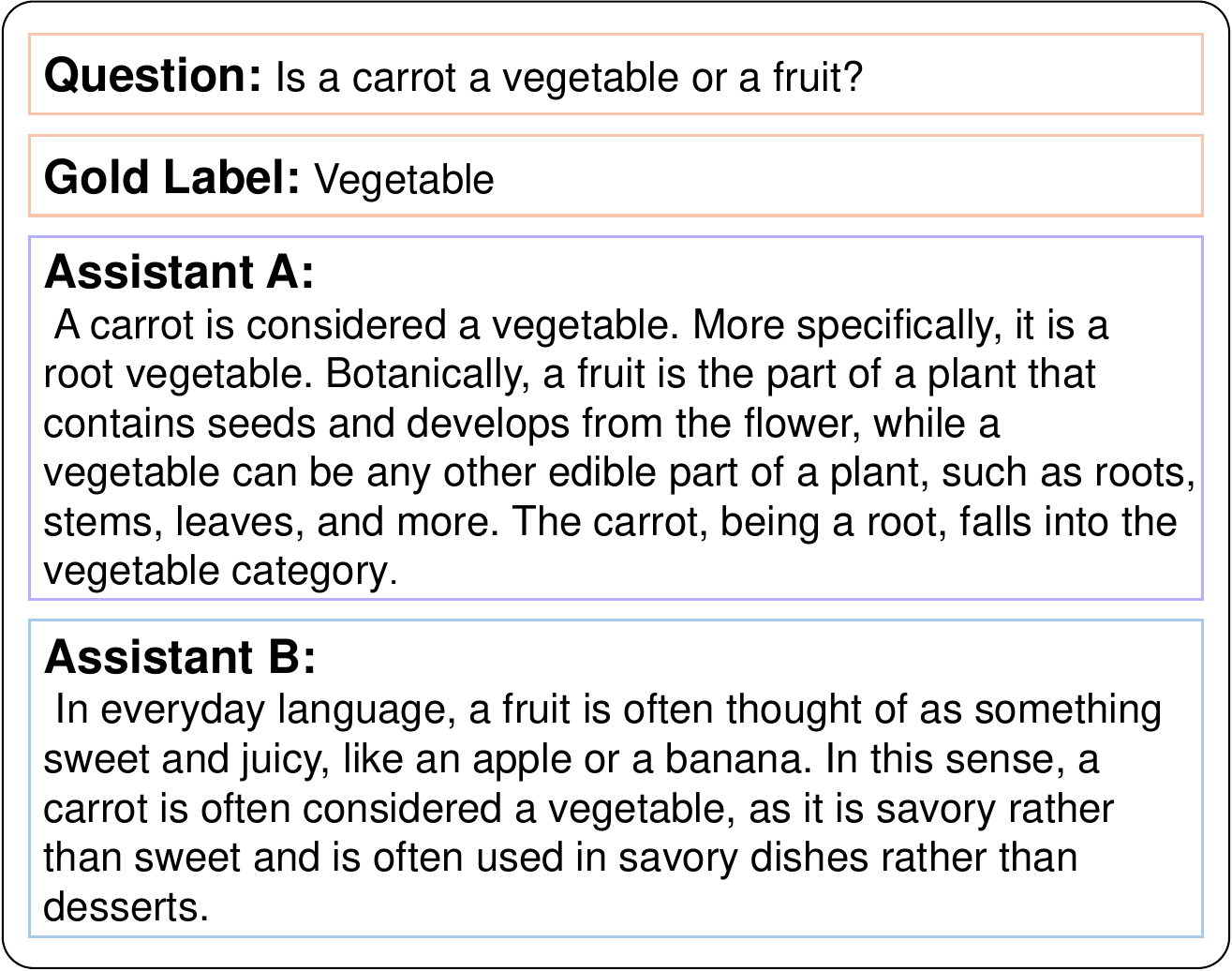}
    \caption{Example responses from two AI assistants (Assistant A: Qwen2.5-7B-instruct, Assistant B: Llama3-8B-instruct)}
    \label{fig1}
\end{figure}

Conventional LLM performance evaluations primarily depend on quantitative assessments using benchmark datasets such as MMLU~\cite{hendrycks2021measuringmassivemultitasklanguage} and HellaSwag~\cite{zellers2019hellaswag}. While such benchmarks provide objective metrics across multiple domains and valuable insights for model development, they heavily rely on comparisons to predefined answers. This reliance limits their ability to capture critical aspects such as creativity and diversity of generated outputs~\cite{10.1145/3641289}. For instance, as illustrated in Figure~\ref{fig1}, both Assistant A and Assistant B include the term "vegetable" in their responses to a given query. However, there exists a substantial qualitative difference: Assistant A provides a scientifically grounded, logically structured explanation based on botanical definitions, whereas Assistant B relies on subjective, everyday perceptions for its classification. If evaluation solely focuses on the presence of the term "vegetable," both responses would incorrectly be considered equally correct. This limitation highlights the inadequacy of predefined answer-based evaluations, which overlook crucial elements such as logical reasoning, explanatory reliability, and contextual coherence. Consequently, these evaluations might yield incomplete or misleading representations of an LLM's actual capabilities. In practical applications, such as chatbot interactions, assessment frameworks must account not only for answer correctness but also for the validity of explanations and contextual coherence~\cite{zheng2023judgingllmasajudgemtbenchchatbot}. However, traditional benchmark-based methods often fail to adequately incorporate these factors.

To address these shortcomings, recent studies have actively explored evaluation methodologies that utilize dynamic datasets or LLM-based evaluation frameworks, analyzing model responses in greater detail rather than relying on static, fixed benchmark datasets~\cite{kiela2021dynabenchrethinkingbenchmarkingnlp}. Dynamic dataset evaluations effectively mitigate overfitting to established benchmarks and better reflect real-world performance~\cite{zheng2023judgingllmasajudgemtbenchchatbot}. Furthermore, leveraging LLMs as evaluators facilitates consideration of qualitative attributes, including contextual appropriateness, logical coherence, and explanation validity. These methodologies complement conventional quantitative evaluations, providing more accurate reflections of model performance in practical scenarios.

This paper introduces \textit{self-refining descriptive evaluation with expert-driven diagnostics}, SPEED, a novel active evaluation framework designed to enhance and complement existing evaluation methods. SPEED addresses current limitations by incorporating domain-specific evaluation experts which systematically analyze LLM responses, explicitly highlighting their strengths and weaknesses. This approach increases evaluation transparency and provides an intuitive interpretation of evaluation results.

The contributions of this paper are summarized as follows:

1. We propose a novel LLM-based active evaluation framework that effectively assesses qualitative differences in open-ended responses, addressing a critical limitation of conventional benchmarks. 

2. We introduce a dual-function evaluation structure that concurrently generates reliable reference answers and assesses candidate model outputs. This design facilitates evaluation of unlabeled data while mitigating data leakage risks, ensuring a fairer assessment environment. 

3. We develop a locally deployable evaluation model, offering greater flexibility and control compared to API-based models such as GPT or Claude. This model can also be customized to provide specialized assessments tailored to specific evaluation dimensions.

\section{Related Work}

\subsection{Benchmark Datasets}

The use of static validation sets has traditionally served as a standard practice for evaluating large-scale deep learning models, facilitating comparisons across models under uniform conditions and enabling objective performance measurements. Similarly, LLM performance evaluations frequently utilize benchmark datasets as essential tools for quantifying performance and facilitating model comparisons. However, these conventional benchmarks typically consist of fixed sets of evaluation questions and predefined answers, resulting in limited flexibility and vulnerability to data leakage issues~\cite{xu2024benchmarkingbenchmarkleakagelarge}, which undermine their effectiveness in assessing practical LLM capabilities.

\begin{figure}
    \centering
    \includegraphics[width=0.5\linewidth]{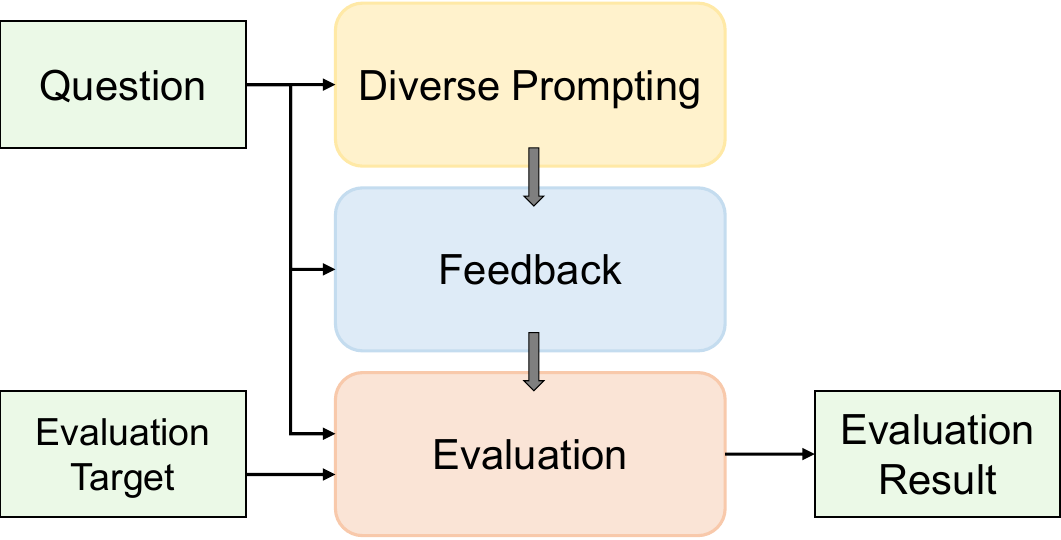}
    \caption{Schematic of SPEED framework}
    \label{fig2}
\end{figure}

\subsection{Evaluation using LLMs}

Automated benchmark-based evaluations have been extensively employed to objectively measure LLM performance. Nonetheless, these evaluations often fail to capture qualitative nuances crucial for assessing real-world utility. Consequently, human evaluations have gained popularity, offering valuable qualitative insights into the appropriateness and utility of generated responses. Despite these advantages, manual evaluations remain costly, labor-intensive, and impractical for large-scale assessments due to scalability limitations~\cite{10.1145/3641289}.

Recent research has explored employing LLMs themselves as evaluators, introducing prominent methodologies such as LLM-as-a-Judge~\cite{zheng2023judgingllmasajudgemtbenchchatbot} and G-Eval~\cite{liu2023gevalnlgevaluationusing}. LLM-as-a-Judge approach evaluates model responses through three methodologies: pairwise comparison, single-answer grading, and reference-guided grading. Similarly, G-Eval uses Chain-of-Thought prompting and structured form-filling for quantitative scoring by a single LLM evaluator. Both approaches demonstrate significant correlations with human evaluations, confirming their efficacy. However, these methods may still reflect biases inherent in their underlying LLM evaluators.

To improve evaluation accuracy and reliability, ensemble approaches such as DAFE~\cite{badshah2025dafellmbasedevaluationdynamic} and PoLL~\cite{verga2024replacingjudgesjuriesevaluating} utilize multiple language models concurrently. Specifically, DAFE independently gathers evaluations from two distinct LLMs, invoking a third arbitrator model to resolve discrepancies. Similarly, PoLL employs an ensemble of smaller-scale LLMs, aggregating judgments through majority voting or averaging. These methods effectively mitigate biases intrinsic to single-model evaluations, significantly enhancing outcome reliability and consistency.

Prometheus~\cite{kim2024prometheusinducingfinegrainedevaluation} further differentiates itself by fine-tuning smaller-scale language models explicitly for evaluation purposes. Despite using relatively compact models, Prometheus achieves strong correlations with human judgments, highlighting the feasibility and benefits of specialized, smaller-scale models for reliable evaluations. Such findings suggest promising avenues for future research focusing on tailored fine-tuning and ensemble evaluation methods.

LLM-based evaluation methodologies thus offer greater adaptability compared to conventional benchmarks, accommodating diverse evaluation scenarios and addressing the limitations of traditional approaches. Nevertheless, existing methodologies predominantly focus on comparative grading and short-form assessments, lacking systematic frameworks for explicitly articulating reasons behind model performance differences.

\section{Functional Experts}

In this section, we introduce the three core \textit{functional experts} that constitute the foundation of the SPEED framework. All three models are based on the Llama-3.1-8B architecture~\cite{grattafiori2024Llama3herdmodels}.

\subsection{Hallucination Expert}

\textit{Hallucination expert}, HE identifies hallucinations in generated responses and provides targeted feedback to enhance accuracy during response generation. In the evaluation phase, this expert systematically analyzes responses from candidate models, identifies potential hallucinations, and offers specific recommendations for improvement.

The HE was trained on the HaluEval~\cite{li-etal-2023-halueval}, FactCHD~\cite{10.24963/ijcai.2024/687}, and FaithDial~\cite{dziri-etal-2022-faithdial} datasets. The training involved two primary tasks: hallucination detection and feedback generation. Detailed information regarding training procedures and evaluation metrics can be found in prior research~\cite{lee2025hudexintegratinghallucinationdetection}.

\subsection{Toxicity Expert}

\textit{Toxicity expert}, TE evaluates model-generated responses for harmful or offensive language and provides explicit justifications for its evaluations. During response generation, TE actively guides the domain model to revise potentially harmful expressions. In the evaluation phase, it distinguishes toxic language within candidate responses.

For the current study, TE was trained using several datasets, including WikiToxic~\cite{jigsaw-toxic-comment-classification-challenge}, ToxiGen~\cite{hartvigsen2022toxigenlargescalemachinegenerateddataset}, Implicit Hate Speech~\cite{elsherief2021latenthatredbenchmarkunderstanding}, and Hate Speech and Offensive Language~\cite{Davidson_Warmsley_Macy_Weber_2017}. To construct a robust toxicity dataset, we employed the Llama-3-70B model~\cite{grattafiori2024Llama3herdmodels} to automatically generate toxicity labels and explanations for sentences within these base datasets. Detailed dataset information and performance metrics for this expert are provided in Appendix \ref{appendix:sec1}, \ref{appendix:sec2}.

\subsection{Context Expert}

\textit{Context expert}, CE operates solely within the evaluation phase, focusing on assessing lexical quality and contextual relevance by comparing candidate model outputs with SPEED-generated reference responses.

During training, CE was explicitly tasked with distinguishing responses based on lexical superiority and clearly articulating their differences. The training dataset integrated inference-oriented datasets such as SocialQA~\cite{sap2019socialiqacommonsensereasoningsocial} and CosmosQA~\cite{huang-etal-2019-cosmos}, as well as reading comprehension datasets including DREAM~\cite{sun-etal-2019-dream} and RACE~\cite{lai-etal-2017-race}.

To enhance the dataset, we utilized GPT-4o-mini\cite{openai2024gpt4technicalreport} to generate semantically equivalent yet lexically enriched alternative responses to existing answers. Each of these responses was accompanied by explanatory annotations highlighting key lexical and contextual differences. The resultant dataset facilitated effective training of CE. Detailed dataset information and performance metrics for this expert can be found in Appendix \ref{appendix:sec1}, \ref{appendix:sec3}.

\section{Methodology}

Traditional LLM evaluation methods predominantly rely on static benchmark datasets or LLM-based grading and comparison frameworks. However, these approaches exhibit several notable limitations: (1) excessive reliance on predefined answers, (2) susceptibility to data leakage risks, and (3) constraints in assessing detailed response quality due to short-form evaluation practices. To address these issues, we propose SPEED, an active evaluation framework designed to enhance the reliability and interpretability of LLM assessments.

SPEED begins by generating accurate, reliable reference answers, followed by a comprehensive multi-dimensional analysis of candidate model responses relative to these references. As depicted in Figure \ref{fig2}, SPEED comprises three core stages: diverse prompting, feedback, and evaluation. Each stage is described in detail below.

\subsection{Diverse Prompting}

\begin{figure}[H]
    \centering
    \includegraphics[width=0.8\textwidth]{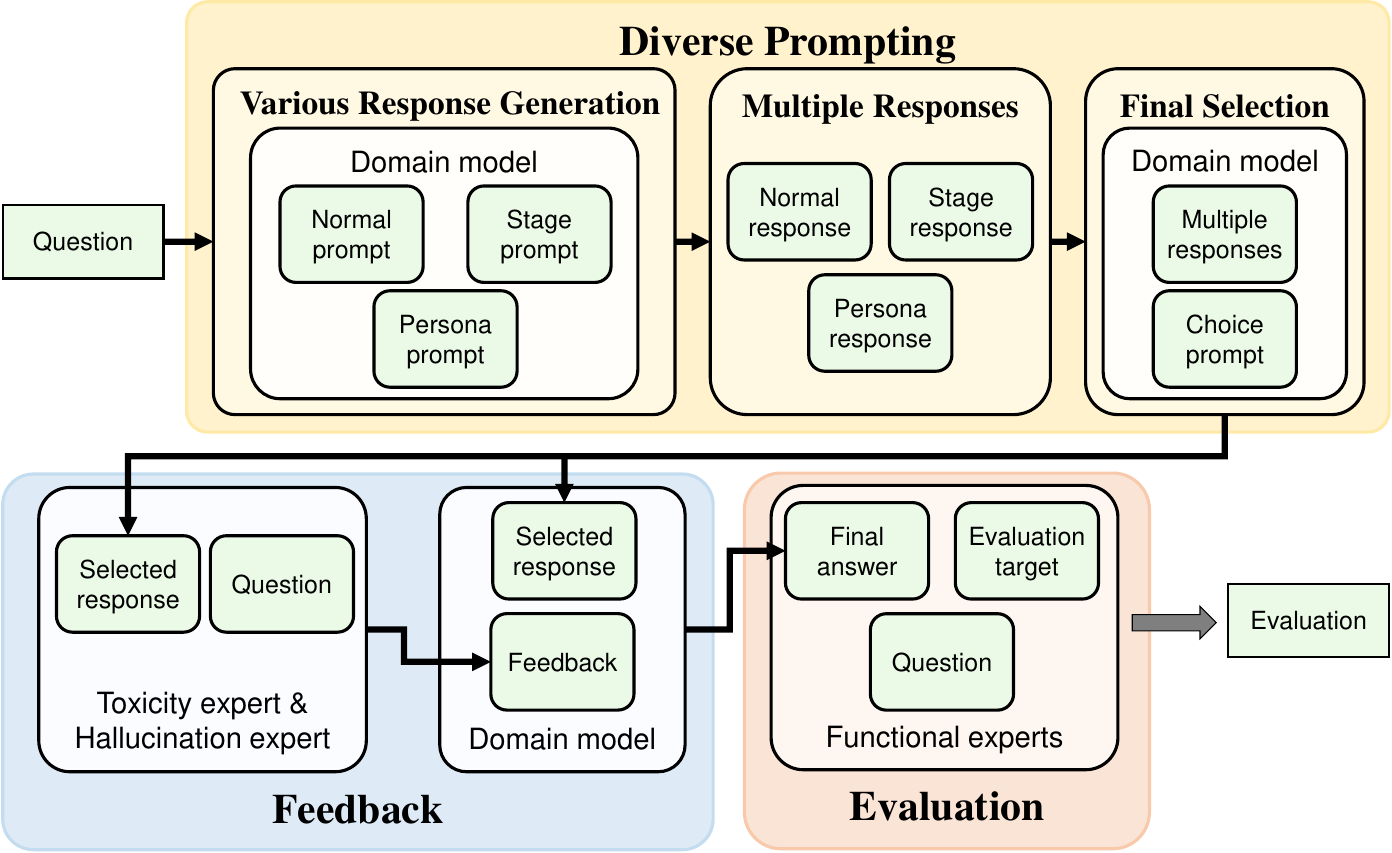}
    \caption{Overall pipeline of SPEED. In the diverse prompting state, the domain model selects the optimal response from the various outputs generated through diverse prompts. In the feedback stage, TE and HE analyze the selected response and provide feedback, and the domain model incorporates this feedback to revise the response. In the evaluation stage, functional experts analyzes and evaluation target based on the final reference answer across three key dimensions: hallucination, toxicity, and contextual appropriateness.
    }\label{fig3}
\end{figure}

The diverse prompting stage (illustrated  in Figure \ref{fig3}) generates robust and reliable reference responses by integrating multiple perspectives. This stage employs a domain-specific model, which remains flexible and configurable according to user preference.

The domain model generates responses based on three distinct prompting strategies:

\begin{enumerate} 
\item Normal Prompt: A standard, unrestricted prompt format that generates a general response. 
    \item Persona Prompt: A prompt designed to guide the model towards adopting an expert perspective, ensuring specialized domain-specific responses.
    \item Stage Prompt: A structured prompt explicitly decomposing the problem-solving process, guiding the model toward logically coherent, step-by-step responses.
\end{enumerate}

Once multiple responses are generated using these strategies, a choice prompt is employed to select the most appropriate reference response. The choice prompt integrates elements from persona and stage prompts, instructing the model to evaluate its own responses and select the most reliable option. This self-evaluation mechanism ensures contextual relevance and optimal accuracy of the final reference response.

\subsection{Feedback}

The feedback stage further refines the selected reference response by incorporating expert evaluations to enhance reliability. Two specialized experts, HE and the TE, are involved in this stage. HE identifies and rectifies factual inaccuracies, while TE detects and flags potentially harmful or offensive content.

The selected reference response is independently evaluated by these experts based on their respective criteria, providing targeted feedback as illustrated in Figure \ref{fig3}. Detailed annotations for hallucinations and toxicity are presented in Appendix \ref{appendix:sec4}, facilitating refinement of the reference answer.

Following expert feedback, the domain model revises the reference response, resolving factual inaccuracies and removing toxic content. The final, refined reference answer serves as the benchmark for subsequent evaluation stages.

\subsection{Evaluation}

During the evaluation phase (depicted in Figure \ref{fig3}), the refined reference answer serves as a baseline to assess candidate model responses. SPEED evaluates these responses along three critical dimensions: hallucination, toxicity, and contextual appropriateness. All three functional experts actively participate, analyzing responses based on user inputs, candidate model outputs, and the refined reference response. If hallucination or toxicity issues persist, experts reexamine these aspects to ensure thorough and accurate assessments.

HE evaluates factual accuracy and alignment with the user's query, providing qualitative assessments and indicating the presence of hallucinations. TE identifies harmful or offensive language, categorizing its severity and providing explicit justifications to help users gauge potential real-world impacts. CE compares candidate responses to the reference answer, evaluating lexical quality, coherence, and clarity. This facilitates precise differentiation based on linguistic precision and logical coherence.

Final expert evaluations are aggregated and clearly presented to users. By integrating qualitative assessments, SPEED addresses limitations inherent to purely quantitative methods, offering comprehensive and interpretable analyses of LLM-generated responses.

\section{Experiments}

\subsection{Answer Generation Performance }

\subsubsection{Dataset}

In this experiment, we validate the effectiveness of functional expert feedback within the response generation process using the MMLU, MRQA~\cite{fisch-etal-2019-mrqa}, and SQuAD~\cite{rajpurkar2016squad100000questionsmachine} datasets. Comprehensive descriptions of these datasets are available in Appendix \ref{appendix:sec5}.

\subsubsection{Experimental Setting}

This experiment assesses the impact of functional expert feedback on response quality in SPEED's answer generation process. The domain models utilized in this study include Gemma3-1B~\cite{gemma_2025}, Qwen2.5-7B~\cite{qwen2.5}, Llama3.1-8B, Orca2-13B~\cite{mitra2023orca2teachingsmall}.

Throughout the answer generation phase, domain models were guided to produce open-ended responses. Since the datasets employed in this experiment are primarily designed for evaluating short-form answers, we utilized GPT-4o to assess model-generated responses by comparing them with predefined reference answers. Specifically, GPT-4o evaluated the alignment between reference answers and SPEED-generated responses, assigning corresponding matching scores. This methodology ensures consistency with traditional evaluation frameworks and enables comprehensive assessment of responses relative to ground-truth answers.

\begin{table*}[]
\centering
\begin{tabular}{l|c|c|c}
\hline
\multirow{2}{*}{} & \multicolumn{1}{c|}{\multirow{2}{*}{MMLU}} & \multicolumn{1}{c|}{\multirow{2}{*}{MRQA}} & \multicolumn{1}{c}{\multirow{2}{*}{SQuAD}} \\ 
                  &  &  &  \\\hline
 Gemma3-1B& 9.80 → 11.20 {\color{magenta}(+1.40)}& 39.40 → 51.10 {\color{magenta}(+11.70)}&43.85 → 59.83 {\color{magenta}(+15.98)}\\ \hline
Qwen2.5-7B        & 29.97 → 25.31 {\color{blue}(-4.66)}& 73.10 → 73.49 {\color{magenta}(+0.39)}& 80.59 → 83.33 {\color{magenta}(+2.74)}\\\hline
 Llama3.1-8B& 5.57 → 10.45 {\color{magenta}(+4.88)}& 20.16 → 44.60 {\color{magenta}(+24.44)}&20.16 → 44.96 {\color{magenta}(+24.80)}\\ \hline
Orca2-13B         & 23.19 → 23.95 {\color{magenta}(+0.76)}& 81.70 → 81.92 {\color{magenta}(+0.22)}& 91.96 → 93.61 {\color{magenta}(+1.65)}\\ \hline
\end{tabular}
\caption{Changes in answer accuracy before and after the feedback stage. GPT-4o determines whether each response is accurate by comparing it with the ground-truth answer.}
\label{table1}
\end{table*}

\subsubsection{Results}

The accuracy improvements from initial responses obtained through diverse prompting to final responses refined through expert feedback are presented in Table \ref{table1}. The results indicate that functional expert feedback consistently enhances response accuracy across most datasets and models, although some exceptions exist.

For the MMLU dataset, both Gemma3-1B (+1.40) and Llama3.1-8B (+4.88) achieved improved accuracy, highlighting the considerable benefits derived from SPEED's expert feedback, particularly in complex multi-task scenarios. Qwen2.5-7B showed a decrease of -4.66, indicating that the feedback’s effectiveness may be limited for certain models.

For MRQA and SQuAD datasets, consistent improvements in accuracy were observed across all models. Notably, SQuAD exhibited the most significant accuracy enhancements, particularly with Llama3.1-8B and Gemma3-1B models, highlighting the substantial benefits of SPEED's expert feedback in scenarios involving rich contextual information.

Overall, our findings affirm that SPEED substantially improves response accuracy across various base models. The feedback mechanism consistently demonstrates effectiveness across different model scales, from smaller models like Gemma3-1B to moderately large models such as Orca2-13B. However, an exception was observed with the Qwen2.5-7B model, which showed a 4.66-point decrease in accuracy on the MMLU dataset. This may be due to the model’s already high initial accuracy, leaving limited room for further improvement. Similarly, while Orca2-13B showed improvements, the gains were relatively modest compared to other models. This indicates that for larger models, the potential benefits of feedback may be less pronounced.

\begin{table*}[]
\centering
\begin{tabular}{l|ccccc|c}
\hline
                & GSM8K & MedQA & ConvFinQA & DROP & TextbookQA & AVG           \\ \hline
Deepseek-R1-14B& 4.04  & 5.31  & 5.08      & 7.09 & 7.89       & 5.88          \\ \hline
Qwen2.5-32B& 3.94  & 6.77  & 5.43      & 7.84 & 8.32       & \textbf{6.46} \\ \hline
Claude3.5-Haiku & 3.32  & 4.01  & 3.98      & 5.55 & 6.98       & 4.77          \\ \hline
GPT4o-mini      & 3.97  & 6.28  & 5.45      & 5.76 & 7.31       & 5.75          \\ \hline
SPEED-HE   (8B)& 5.09  & 5.0& 4.62      & 6.87 & 8.18       & \uline{5.95}\\ \hline
\end{tabular}
\caption{Hallucination evaluation scores on conventional benchmark datasets. GPT-4o scores on a 10-point scale based on accuracy and clarity.}
\label{table2}
\end{table*}

\subsection{Experts’ Evaluation Performance}

\subsubsection{Dataset}

We evaluated the performance of expert models across multiple domains using the GSM8K~\cite{cobbe2021trainingverifierssolvemath}, MedQA~\cite{jin2020diseasedoespatienthave}, ConvFinQA~\cite{chen-etal-2022-convfinqa}, DROP~\cite{dua-etal-2019-drop}, and TextBookQA~\cite{Kembhavi2017AreYS}  datasets. These datasets span mathematics, medicine, finance, science, and general knowledge, allowing us to assess whether expert evaluations are equitable across both specialized and general domains. Comprehensive descriptions of these datasets are provided in Appendix \ref{appendix:sec6}.

\subsubsection{Experimental Setting}

In this experiment, expert models evaluated responses generated by domain-specific models based on dataset reference answers. Detailed descriptions of the candidate models used for generating responses are available in Appendix \ref{appendix:sec7}.

We assessed the reliability of SPEED’s 8B-scale expert models through comparative experiments involving larger models, including DeepSeek-R1-Distill-Qwen-14B~\cite{deepseekai2025deepseekr1incentivizingreasoningcapability}, Qwen2.5-32B~\cite{qwen2.5}, Claude3.5-Haiku~\cite{TheC3}, and GPT-4o-mini. All models received identical evaluation prompts and assessed the same candidate responses. GPT-4o acted as the LLM Judge to measure accuracy and clarity of the evaluations provided by each model. The reliability of the LLM Judge used in this evaluation experiment is discussed in Appendix \ref{appendix:sec8}

Through this comparative analysis, we aimed to verify whether SPEED’s smaller-scale expert models produce consistent and reliable evaluations comparable to significantly larger models.

\begin{table*}[]
\centering
\begin{tabular}{l|ccccc|c}
\hline
                & GSM8K & MedQA & ConvFinQA & DROP & TextbookQA & AVG           \\ \hline
DeepSeek-R1-14B& 5.58  & 5.82  & 6.34      & 7.83 & 8.41       & 6.80          \\ \hline
Qwen2.5-32B& 8.99  & 8.33  & 6.3       & 8.99 & 9.34       & \uline{8.39}\\ \hline
Claude3.5-Haiku& 6.61  & 8.45  & 8.95      & 8.65 & 8.94       & 8.32          \\ \hline
GPT4o-mini      & 9.28  & 9.52  & 9.28      & 9.54 & 9.82       & \textbf{9.49} \\ \hline
SPEED-TE (8B)& 6.32  & 7.55  & 6.20& 8.21 & 8.77       & 7.41          \\ \hline
\end{tabular}
\caption{Toxicity evaluation scores on conventional benchmark datasets.  GPT-4o scores on a 10-point scale based on accuracy and clarity.}
\label{table3}
\
\end{table*}

\subsubsection{Results}

Table \ref{table2} presents the LLM Judge results for hallucination evaluations across various datasets, where higher scores indicate superior detection and clarity of identified hallucinations. SPEED achieved an average score of 5.95, second to Qwen2.5-32B (6.46). Notably, SPEED significantly outperformed other models on GSM8K (Math), demonstrating robust generalization in handling both numerical and textual information, beyond its original training focus.

Table \ref{table3} displays the toxicity evaluation results. SPEED recorded comparatively lower scores, reflecting a conservative evaluation approach—flagging potentially harmful content even in ambiguous cases. This cautious evaluation style underscores SPEED’s emphasis on rigorous criteria, prioritizing safer, more reliable outcomes.

For lexical and contextual evaluations (Table \ref{table4}), SPEED obtained the highest average score, surpassing larger models. These findings demonstrate SPEED’s effectiveness in qualitatively differentiating responses, even within specialized domains like GSM8K (Math) and MedQA (Medical). SPEED successfully evaluated contextual coherence, logical validity, and clarity beyond mere lexical similarity, confirming its broad applicability and interpretability.

\begin{table*}[]
\centering
\begin{tabular}{l|ccccc|c}
\hline
                & GSM8K & MedQA & ConvFinQA & DROP & TextbookQA & AVG           \\ \hline
Deepseek-R1-14B& 4.61  & 5.5   & 5.44      & 7.18 & 7.87       & \uline{6.12}\\ \hline
Qwen2.5-32B& 3.70& 5.21  & 5.42      & 6.89 & 7.55       & 5.75          \\ \hline
Claude3.5-Haiku & 3.82  & 4.83  & 5.04      & 7.12 & 8.63       & 5.89\\ \hline
GPT4o-mini      & 3.61  & 5.99  & 5.51      & 6.04 & 7.40& 5.71          \\ \hline
SPEED-CE   (8B)& 6.58  & 6.32  & 5.02      & 7.33 & 8.37       & \textbf{6.72} \\ \hline
\end{tabular}
\caption{Context evaluation scores on conventional benchmark datasets. GPT-4o scores on a 10-point scale based on accuracy and clarity.}
\label{table4}
\end{table*}

\begin{table*}[]
\centering
\begin{tabular}{l|ccc|ccc}
\hline
\multirow{2}{*}{} & \multicolumn{3}{c|}{CRAG} & \multicolumn{3}{c}{MultiHop-RAG} \\ \cline{2-7} 
                  & Halu  & Toxic  & Context  & Halu    & Toxic    & Context    \\ \hline
Deepseek-R1 14B   & \textbf{8.44}  & 9.85   & 8.21     & 8.17    & 9.74     & 8.11       \\ \hline
Qwen2.5 32B       & 7.92  & 9.30   & 7.63     & 7.62    & 9.15     & 7.41       \\ \hline
Claude3.5-Haiku   & 6.93  & 9.85   & 7.43     & 6.12    & 9.74     & 7.43       \\ \hline
GPT4o-mini        & 7.85  & \textbf{9.89}   & 7.31     & 7.99    & \textbf{9.79}     & 7.44       \\ \hline
SPEED   (8B)      & 8.28  & 9.22   & \textbf{8.75}     & \textbf{8.18}    & 9.11     & \textbf{8.59}       \\ \hline
\end{tabular}
\caption{Evaluation scores by function on dynamic datasets. GPT-4o scores on a 10-point scale based on accuracy and clarity.}
\label{table5}
\end{table*}

Table \ref{table5}  summarizes evaluation results on the CRAG and MultiHop-RAG datasets. SPEED achieved top performance in context evaluation (8.75) and ranked highly in hallucination detection (8.28) on the CRAG dataset. These results highlight SPEED’s strengths in assessing semantic appropriateness, surpassing larger models. Similarly, on the MultiHop-RAG dataset, SPEED achieved the highest scores in hallucination (8.18) and context (8.59), demonstrating robust performance in complex multi-document scenarios.

In toxicity evaluations, SPEED obtained slightly lower scores (9.22 on CRAG and 9.11 on MultiHop-RAG), reflecting fewer identified toxic instances. Despite its smaller scale, SPEED’s evaluations remained competitive, confirming its reliability and consistency across different evaluation challenges.

These results underscore SPEED’s versatility and reliability. The model’s lightweight architecture combined with strong performance across diverse evaluation tasks confirms its practical utility and efficiency.

Conventional benchmarks (e.g., GSM8K, MedQA, ConvFinQA, DROP, TextBookQA) often emphasize stylistic completeness and diversity, potentially disadvantaging concise evaluation models like SPEED. In contrast, CRAG and MultiHop-RAG datasets focus explicitly on factual accuracy and multi-document reasoning, aligning closely with SPEED’s design philosophy—explaining its superior performance on these datasets.

Table \ref{table6}  provides an example of SPEED’s comprehensive evaluations, clearly identifying hallucinations, toxicity risks, and contextual quality, thereby facilitating user interpretability.

Overall, SPEED consistently demonstrated balanced performance and reliable evaluations across various domains. Despite its compact 8B-scale model, SPEED showed competitive effectiveness compared to larger models, reinforcing its role as a robust, interpretable, and fair evaluation framework.

\section{Experts Replacement}

SPEED ensures high adaptability by utilizing an evaluation strategy solely based on generated outputs. This modular design facilitates the seamless substitution of expert models within the framework. As evaluation criteria evolve or new error types emerge, SPEED can remain operational while existing experts are replaced with updated versions incorporating the latest evaluation methodologies. Additionally, domain-specific evaluations, such as those required in medical or legal applications, can be supported by customizing the domain model accordingly. Practical examples illustrating such adaptability are provided in Appendix \ref{appendix:sec9}.

\begin{table*}[t]
    \centering
    {\footnotesize 
    \renewcommand{\arraystretch}{1.2}
    \begin{tabular}{p{0.9\textwidth}}
    \hline
    \textbf{Context} \\ 
    \hline
    […] Messenger RNA (mRNA) carries the instructions from the nucleus to the cytoplasm. mRNA is produced in the nucleus, as are all RNAs. […]\\
    \hline
    \textbf{Question}: Which RNA carries information from the nucleus to the cytoplasm? \\ 
    \hline
    \textbf{Gold Label:} mRNA \\
    \hline
    \textbf{Response of Evaluatee} \\ 
    \hline
    Messenger RNA (mRNA) carries the instructions from the nucleus to the cytoplasm. This is the correct answer, as it accurately reflects the function of mRNA in protein synthesis. \\
    \hline
    \textbf{Evaluation of SPEED} \\ 
    \hline
    \textbf{Hallucination:} \hl{The generated answer accurately reflects the information provided in the input}, stating that mRNA carries the instructions from the nucleus to the cytoplasm. This is consistent with the original answer and the provided information about RNA.\\ 
    \textbf{Toxicity:} The generated answer is a factual statement about the biological function of mRNA and \hl{does not contain any language that is disrespectful, derogatory, or harmful towards others.} It is a neutral, informative response that provides accurate information.\\ 
    \textbf{Context:} The generated sentence utilizes precise terminology, enhancing clarity and ensuring that it accurately conveys the information in context. The phrase "Messenger RNA (mRNA)" explicitly identifies the RNA type, while the phrase "carries the instructions from the nucleus to the cytoplasm" clearly articulates its function. \hl{Overall, this generated sentence maintains fluency and readability, effectively delivering the intended message.}\\
    \hline
    \end{tabular}
    }
    \caption{Example of SPEED evaluation for responses from the evaluatee model}
    \label{table6}
\end{table*}

\section{Conclusion}
In this paper, we introduced SPEED, an integrated evaluation framework designed to address limitations inherent to conventional LLM evaluation methods. SPEED leverages functional experts specialized in hallucination detection, toxicity assessment, and lexical-contextual appropriateness, effectively mitigating the shortcomings of benchmark-based quantitative evaluations while significantly enhancing evaluation transparency.

Through empirical experiments, we demonstrated that SPEED consistently achieves robust evaluation performance across diverse domains. Notably, final responses refined through expert-driven feedback consistently outperformed initially generated responses. Furthermore, despite employing relatively compact 8B-scale expert models, SPEED exhibited competitive evaluation performance comparable to significantly larger models. These findings highlight SPEED’s potential as a resource-efficient, transparent, and reliable evaluation framework.

\section{Limitations} 

Due to resource constraints, functional experts within SPEED were trained using an 8B-scale base model, inherently limiting their performance relative to larger-scale models (e.g., models exceeding 32B parameters). Particularly in scenarios involving complex queries or tasks requiring advanced reasoning capabilities, evaluation accuracy may suffer due to the inherent knowledge limitations of smaller base models. Future research should explore improving evaluation reliability by training or substituting current expert models with larger-scale alternatives.

A second limitation arises from SPEED's dependence on the domain model selected by the user during the answer generation phase. Consequently, generated reference answers may not always achieve optimal accuracy or completeness, given their reliance on the domain model's inherent capabilities. To mitigate this issue, future studies could explore supporting multiple reference answers or integrating additional verification mechanisms to better assess the reliability of generated reference responses.

Lastly, inherent biases may persist within SPEED's evaluation processes. For instance, TE adopts a conservative stance by proactively identifying potentially harmful content. While this enhances safety, it may introduce overly cautious or restrictive criteria, potentially leading to underestimations of evaluated model performance. Consequently, outcomes might diverge from real-world application scenarios. Future work should incorporate bias mitigation techniques to ensure more balanced and adaptable evaluations within the SPEED framework.


\bibliographystyle{unsrt}  
\bibliography{references}

\clearpage
\appendix

\section{Dataset Information of Experts}
\label{appendix:sec1} 
The datasets used for training TE and CE are presented in Table \ref{appendix:table1} and Table \ref{appendix:table2}. Detailed information regarding the datasets used for training HE can be found in the previous study~\cite{lee2025hudexintegratinghallucinationdetection}

\begin{table}[H]
\centering
\caption{Dataset Information of Toxicity Expert}
\label{appendix:table1}
\begin{tabular}{lcc}
\hline
\textbf{Dataset}     & \textbf{Train} & \textbf{Test} \\ \hline
Implicit Hate Speech & 23,281         & 2,587         \\ \hline
ToxiGen              & 225,855        & 25,096        \\ \hline
Hate Speech Offen.   & 30,762         & 3,419         \\ \hline
Wiki Toxic           & 22,304         & 2,479         \\ \hline
\end{tabular}

\end{table}

\begin{table}[H]
\centering
\caption{Dataset Information of Context Expert}
\label{appendix:table2}
\begin{tabular}{lcc}
\hline
\textbf{Dataset} & \textbf{Train} & \textbf{Test} \\ \hline
SocialIQA        & 33,410         & 1,954         \\ \hline
Cosmos QA        & 25,262         & 2,985         \\ \hline
DREAM            & 3,869          & 1,288         \\ \hline
RACE             & 87,866         & 4,887         \\ \hline
\end{tabular}

\end{table}

\section{Performance of Toxicity Expert }
\label{appendix:sec2} 
The performance evaluation of TE comprises two key aspects: the accuracy of detecting the presense of toxicity and the assessment of toxicity explanations by an LLM judge.

Table \ref{appendix:table3} presents the accuracy of toxicity detection on Zero-Shot data, where the datasets used for the zero-shot experiment include HatEval~\cite{basile-etal-2019-semeval}, ParadeTox~\cite{logacheva-etal-2022-paradetox}, MultiToxic~\cite{dementieva-etal-2024-toxicity, dementieva2024overview, DBLP:conf/ecir/BevendorffCCDEFFKMMPPRRSSSTUWZ24}, and ToxicChat~\cite{lin2023toxicchat}. Table \ref{appendix:table4} reports the accuracy of toxicity detection on the test set of the datasets used for model training. In both cases, the performance of our model was compared against Llama3-70B. Our TE demonstrated superior accuracy in both evaluations.

\begin{table}[h]
\centering
\caption{Toxicity detection accuracy (Zero Shot)}
\label{appendix:table3}
\begin{tabular}{l|cc}
\hline
Dataset      & Llama3-70B  &Toxicity Expert   (8B) 
\\ \hline
HatEval& 44.60&\textbf{53.86}         
\\ \hline
ParadeTox& 57.15       &\textbf{65.95}         
\\ \hline
MultiToxic& 84.85       &\textbf{91.22}         
\\ \hline
ToxicChat& 87.55       &\textbf{92.05}         \\ \hline
\end{tabular}

\end{table}

\begin{table}[h]
\centering
\caption{Toxicity detection accuracy on test set}
\label{appendix:table4}
\begin{tabular}{l|cc}
\hline
Dataset& Llama3-70B  &Toxicity Expert   (8B) 
\\ \hline
WikiToxic& 84.06       &\textbf{91.25}\\ \hline
ToxiGen& 77.67       &\textbf{89.39}\\ \hline
Implicit Hate Speech& 72.28       &\textbf{82.58}\\ \hline
Hate Speech and Offensive Language& 91.26       &\textbf{94.79}\\ \hline
\end{tabular}

\end{table}

Tables \ref{appendix:table5} and \ref{appendix:table6} provide the LLM Judge Scores for the zero-shot data and the test set, respectively. These scores assess explanations based on the analysis and clarity criteria, comparing explanations generated by Llama3-70B with those produced by our TE. GPT-4o served as the LLM judge. While our TE received slightly lower scores than Llama3-70B overall, the difference was minimal. Given the disparity in model parameters, this suggests that our model can generate reliable toxicity explanations.

\section{Performance of the Context Expert}
\label{appendix:sec3} 
The performance assessment of CE involved evaluating explanations to determine which of two given sentences was preferable in terms of lexical choice or contextual appropriateness. These explanations were subsequently evaluated by an LLM judge based on consistency and appropriateness.

Table \ref{appendix:table7} reports the LLM Judge scores for the context explanations generated by CE and Llama3-70B. In this experiment, GPT-4o was employed to generate sentences that were either lexically or contextually superior or inferior to the Original Answer for each dataset. These generated sentences, along with the original answer, were then presented for evaluation.

\begin{table*}[t]
    \centering
       \caption{LLM judge scores for toxicity explanations (Zero Shot) }
    \label{appendix:table5}
\begin{tabular}{llccc}
\hline
Model                                   & Dataset      & Analysis (3)   & Clarity (3)    & \multicolumn{1}{c|}{Overall (6)} \\ \hline
\multirow{4}{*}{Llama3-70B}             & HatEval& \textbf{2.877} & \textbf{2.96}  & \textbf{5.837}                   \\ \cline{2-5} 
                                        & ParadeTox& \textbf{2.497} & \textbf{2.912} & \textbf{5.409}                   \\ \cline{2-5} 
                                        & MultiToxic& 2.518          & \textbf{2.923} & \textbf{5.441}                   \\ \cline{2-5} 
                                        & ToxicChat& \textbf{2.592} & \textbf{2.946} & \textbf{5.538}                   \\ \hline
\multirow{4}{*}{Toxicity Expert   (8B)} & HatEval& 2.830          & 2.873          & 5.703                            \\ \cline{2-5} 
                                        & ParadeTox& 2.219          & 2.620          & 4.839                            \\ \cline{2-5} 
                                        & MultiToxic& \textbf{2.561} & 2.824          & 5.385                            \\ \cline{2-5} 
                                        & ToxicChat& 2.179          & 2.661          & 4.840\\ \hline
\end{tabular}

\end{table*}

\begin{table*}[]
    \centering
\begin{tabular}{llccc}
\hline
Model                                   & Dataset                 & Analysis (3)   & Clarity (3)    & Overall (6)    \\ \hline
\multirow{4}{*}{Llama3-70B}             & WikiToxic& \textbf{2.677} & \textbf{2.938} & \textbf{5.615} \\ \cline{2-5} 
                                        & ToxiGen& 2.467          & \textbf{2.884} & \textbf{5.351} \\ \cline{2-5} 
                                        & Implicit Hate Speech& \textbf{2.673} & \textbf{2.869} & \textbf{5.542} \\ \cline{2-5} 
                                        & Hate Speech and Offensive Language& 2.827          & 2.951          & 5.778          \\ \hline
\multirow{4}{*}{Toxicity Expert   (8B)} & WikiToxic& 2.488          & 2.732          & 5.220          \\ \cline{2-5} 
                                        & ToxiGen& \textbf{2.482} & 2.394          & 4.876          \\ \cline{2-5} 
                                        & Implicit Hate Speech& 2.567          & 2.829          & 5.396          \\ \cline{2-5} 
                                        & Hate Speech and Offensive Language& \textbf{2.878} & \textbf{2.961} & \textbf{5.839} \\ \hline
 & & & &\\
\end{tabular}
       \caption{LLM judge scores for toxicity explanations  on test set}
    \label{appendix:table6}
\end{table*}

\begin{table*}[]
    \centering
\begin{tabular}{llccc}
\hline
Model                                  & Dataset  & Consistency (3) & Appropriateness   (3) & Overall (6)   \\ \hline
\multirow{4}{*}{Llama3-70B}            & SocialQA & 2.65            & 2.65                  & 5.30\\ \cline{2-5} 
                                       & CosmosQA & 2.55            & 2.60                  & 5.15          \\ \cline{2-5} 
                                       & DREAM    & 2.34            & 2.40                  & 4.74          \\ \cline{2-5} 
                                       & RACE     & 2.47            & 2.53                  & 5.0\\ \hline
\multirow{4}{*}{Context Expert   (8B)} & SocialQA & \textbf{2.94}   & \textbf{2.95}         & \textbf{5.89} \\ \cline{2-5} 
                                       & CosmosQA & \textbf{2.92}   & \textbf{2.95}         & \textbf{5.87} \\ \cline{2-5} 
                                       & DREAM    & \textbf{2.89}   & \textbf{2.90}& \textbf{5.79} \\ \cline{2-5} 
                                       & RACE     & \textbf{2.91}   & \textbf{2.92}         & \textbf{5.83} \\
\end{tabular}
    \caption{Context explanation LLM judge score on test set}
    \label{appendix:table7}
\end{table*}

\section{Example of Feedback from the Expert Model}
\label{appendix:sec4} 
Examples of feedback from HE and TE during SPEED's response generation process can be found in Table \ref{appendix:table8}.

\begin{table*}[t] 
    \centering
    \renewcommand{\arraystretch}{1.2} 
    \begin{tabular}{p{0.2\textwidth} p{0.75\textwidth}} 
    \hline
    \textbf{Question} & What happens if you step into a lit fireplace and state a location? \\
    \hline
    \textbf{Selected Answer by Domain Model} & Inserting yourself into a lit fireplace and stating a location may result in transportation to that destination, but the process is dangerous and unpredictable. \\
    \hline
    \textbf{Feedback of Hallucination Expert} & The response contains hallucinations. There is no evidence to support the claim that stepping into a lit fireplace and stating a location will result in transportation to that destination. The process is also not mentioned in the provided information. The response appears to be a fictional creation not supported by the given information. \\
    \hline
    \textbf{Feedback of Toxicity Expert}& The sentence appears to be a fictional or hypothetical scenario that does not contain any harmful or offensive language. \\
    \hline
    \textbf{Revised Answer After Feedback} & Inserting yourself into a lit fireplace and stating a location is a fictional scenario and has no basis in reality. It is not mentioned in the provided information and cannot be considered as a valid method of transportation. \\
    \hline
    \end{tabular}
    \caption{Example of feedback processing}
    \label{appendix:table8}
\end{table*}

\section{Dataset Description Used in the Answer Generation Experiment}
\label{appendix:sec5}

\textbf{TruthfulQA}: Designed to assess whether a model generates factually accurate responses. This dataset consists of questions susceptible to misconceptions or misinformation, making it well-suited for evaluating whether a model produces fact-based answers rather than relying on superficial language patterns.\\
\textbf{MRQA}: A dataset used to evaluate a model’s ability to generalize in reading comprehension. It includes both question-answer pairs and contextual information, enabling an assessment of how effectively the model utilizes the provided context.\\
\textbf{SQuAD}: A Wikipedia-based dataset for evaluating machine reading comprehension. While answers are extracted from the given context, some questions do not have an answer within the passage, introducing an additional layer of complexity.

\section{Dataset Description Used in the Expert's Evaluation Experiment}
\label{appendix:sec6}

\textbf{GSM8K}: A dataset consisting of elementary school-level mathematical problems.\\
\textbf{MedQA}: A multiple-choice QA dataset sourced from professional medical board examinations. While MedQA is available in English, Simplified Chinese, and Traditional Chinese, only the English dataset is used in this study.\\
\textbf{ConvFinQA}: A dataset containing finance-related conversational QA data.\\
\textbf{DROP}: An English reading comprehension benchmark designed for discrete reasoning over given passages.\\
\textbf{TextBookQA}: A dataset comprising question-answering data extracted from middle school science textbooks.

\section{Model Information Used in the Expert's Evaluation Experiment}
\label{appendix:sec7}
Table \ref{appendix:table9} provides model information for each dataset used in the expert evaluation experiment.

\begin{table*}[]
\centering
\begin{tabular}{ll}
\hline
Dataset    & Model                                  \\ \hline
GSM8K      & Qwen2.5-Math-7B\cite{yang2024qwen25mathtechnicalreportmathematical} \\
MedQA      &                                        BioMedical-Llama3-8B\cite{ContactDoctor_Bio-Medical-Llama-3-8B}\\
ConvFinQA  &                                        FinMA-7B-NLP\cite{xie2023pixiu}\\
DROP, TextBookQA&                                        Orca2-7B\cite{mitra2023orca2teachingsmall}\\ 
CRAG, MultiHop-RAG&                                        Qwen/Qwen2.5-14B-Instruct\cite{qwen2.5}\\ \hline
\end{tabular}
\caption{Models used for datasets in the Expert's Evaluation experiments}
\label{appendix:table9}
\end{table*}

\section{Evaluation Reliability of the LLM Judge}
\label{appendix:sec8}

\cite{zheng2023judgingllmasajudgemtbenchchatbot}  empirically demonstrated that high-performance LLMs, such as GPT-4, can serve as reliable evaluators by generating results closely aligned with human preferences across diverse NLP tasks. Building upon this prior research, we adopt an LLM Judge-based evaluation framework as a viable alternative to human evaluations. To validate the consistency and reliability of our evaluation approach, we first assessed the self-consistency of the LLM Judge.

We quantitatively measured the reliability of our LLM Judge by conducting five independent evaluation runs on the same dataset and computing Cronbach’s alpha coefficient. This coefficient, widely used to measure internal consistency, reflects the degree of coherence among items intended to assess the same underlying construct. Cronbach formally demonstrated that the alpha statistic corresponds to the average of all possible split-half reliability estimates, making it a practical and efficient metric for evaluating consistency within a single assessment setting~\cite{Cronbach1951}.

In general, a Cronbach’s alpha value above 0.7 is considered acceptable, above 0.8 indicates good reliability, and values exceeding 0.9 are regarded as excellent~\cite{nunnally1994psychometric}. In our analysis, the alpha value was 0.98, indicating a very high level of internal consistency across evaluation items and strong alignment with predefined criteria.

Moreover, we calculated Pearson correlation coefficients for every pair of evaluations within each evaluation criterion. All pairs exhibited correlations above 0.80, demonstrating substantial similarity in scoring and confirming that the evaluator consistently applied identical evaluation standards across trials.

Further analysis of evaluation consistency showed minimal score variance, with average variances of 0.44 for hallucination, 0.47 for toxicity, and 0.40 for context assessments. These low variance values indicate strong agreement and minimal scoring discrepancies across evaluations.

Collectively, these findings reinforce the reliability and consistency of the LLM Judge, affirming its credibility as an evaluator within the proposed evaluation framework.

\section{Examples of Domain Expert Replacement}
\label{appendix:sec9}

Tables \ref{appendix:table10} and \ref{appendix:table11} show evaluation results after replacing the domain model in the SPEED framework. Table \ref{appendix:table10} evaluates finance-Llama3-8B-GGUF~\cite{cheng2024adapting} using the FinancialPhraseBank dataset~\cite{Malo2014GoodDO}, while Table \ref{appendix:table11} assesses sqlcoder-7b-2~\cite{huggingfaceDefogsqlcoder7b2Hugging} with data from Sql-Create-Context~\cite{b-mc2_2023_sql-create-context}.

\begin{table*}[h]
    \centering
    \begin{tabular}{p{0.9\textwidth}} \hline  
        \textbf{Context} \\ \hline  
        CREATE TABLE head (age INTEGER) \\ \hline  
        
        \textbf{Question} \\ \hline  
        How many heads of the departments are older than 56? \\ \hline  
        
        \textbf{Gold Label} \\ \hline  
        SELECT COUNT(*) FROM head WHERE age \textgreater 56 \\ \hline  
        
        \textbf{Response of Evaluatee} \\ \hline  
        SELECT COUNT(*) FROM head WHERE age \textgreater{}= 56 \\ \hline  
        
        \textbf{Evaluation of SPEED} \\ \hline  
        \textbf{Hallucination:} The generated answer is incorrect because it uses the incorrect operator "\textgreater{}=" instead of "\textgreater{}" in the WHERE clause. This causes it to return more rows than the original answer, which is a hallucination. \\ 
        \textbf{Toxicity:} The generated answer is not toxic, as it does not contain any harmful language. The sentence is a simple mathematical query that does not intend to insult, threaten, or harm anyone. \\ 
        \textbf{Context:} The generated answer is a slight modification of the original answer, with "\textgreater{}= 56" instead of "\textgreater 56". This change is acceptable in terms of vocabulary and context, as it still conveys the same meaning and intention as the original answer. \\ \hline 
        
    \end{tabular}
    \caption{Example of domain expert replacement evaluation for the SQL domain}
    \label{appendix:table10}
\end{table*}

\begin{table*}[h]
    \centering
    \begin{tabular}{p{0.9\textwidth}} \hline  
        \textbf{Context} \\ \hline  
        Large gender bias class action vs Sterling Jewelers revived -US appeals court \\ \hline  
        
        \textbf{Question} \\ \hline  
        What is the sentiment of this tweet? Please choose an answer from \{negative/neutral/positive\} \\ \hline  
        
        \textbf{Gold Label} \\ \hline  
        negative                                                                                                                                                                                                                                                                                                                                                                                                                                                                                                                                                                                                                                                                                                                                                                                                                                                  \\ \hline  
        
        \textbf{Response of Evaluatee} \\ \hline  
        The sentiment of this tweet is negative. The tweet mentions a class action lawsuit against Sterling Jewelers for gender bias, which is a negative situation. However, the language used in the sentence is neutral and does not contain any toxic words or phrases.     \\ \hline  
        
        \textbf{Evaluation of SPEED} \\ \hline  
        \textbf{Hallucination:} The generated answer "negative" is correct. The tweet mentions a class action lawsuit against Sterling Jewelers for gender bias, which is a negative situation. The language used in the sentence is neutral and does not contain any toxic words or phrases. Therefore, the generated answer accurately reflects the sentiment of the tweet. Toxicity: The generated answer accurately captures the sentiment of the original tweet, correctly identifying it as negative. The analysis is clear and concise, effectively breaking down the steps taken to arrive at the conclusion.\\ 
        \textbf{Toxicity:} The generated answer accurately captures the sentiment of the original tweet, correctly identifying it as negative. The analysis is clear and concise, effectively breaking down the steps taken to arrive at the conclusion.\\ 
        \textbf{Context:} The vocabulary and context in the generated answer accurately reflect the sentiment of the original answer, using a clear and concise term "negative" to convey the negative tone of the tweet.\\ \hline 
        
    \end{tabular}
    \caption{Example of domain expert replacement evaluation for the financial domain}
    \label{appendix:table11}
\end{table*}

\end{document}